# Hybrid Deep Learning Modeling Approach to Predict Natural Gas Consumption of Home Subscribers on Limited Data

Milad Firoozeh, Nader Dashti*, Mohammad Ali Hatefi

Department of Economics and Management, Petroleum University of Technology (PUT), Tehran, Iran

A B S T R A C T

Today, natural gas, as a clean fuel and the best alternative to crude oil, covers a significant part of global demand. Iran is one of the largest countries with energy resources and in terms of gas is the second-largest country in the world. But, due to the increase in population and energy consumption, it faces problems such as pressure drops and gas outages yearly in cold seasons and therefore it is necessary to control gas consumption, especially in the residential sector, which has the largest share in Iran. This study aims to analyze and predict gas consumption for residential customers in Zanjan province, Iran, using machine learning models, including LSTM, GRU, and a hybrid BiLSTM-XGBoost model. The dataset consists of gas consumption and meteorology data collected over six years, from 2017 to 2022. The models were trained and evaluated based on their ability to accurately predict consumption patterns.

The results indicate that the hybrid BiLSTM-XGBoost model outperformed the other models in terms of accuracy, with lower Root Mean Squared Error (RMSE), Mean Absolute Percentage Error (MAPE) values, and Mean Percentage Error (MPE). Additionally, the Hybrid model demonstrated robust performance, particularly in scenarios with limited data. The findings suggest that machine learning approaches, particularly hybrid models, can be effectively utilized to manage and predict gas consumption, contributing to more efficient resource management and reducing seasonal shortages. This study highlights the importance of incorporating geographical and climatic factors in predictive modeling, as these significantly influence gas usage across different regions.

## 1. Introduction

Energy has long been considered one of the essential human needs. Humans met their needs using wind and water energies. Over time, with the discovery and extraction of coal, a new era emerged to develop industry and transportation. The exploration and production of oil, entered cheaper and more functional energy in the world (Soldo, 2012). In 1785, natural gas was first commercially used in Britain that produced from coal, and to this day, this energy source has played a significant

---

* Corresponding author.
E-mail addresses: miladfiroozeh@outlook.com (M. Firoozeh), dashti_n@put.ac.ir (N. Dashti), hatefi@put.ac.ir (M.A.Hatefi)

role in most parts of human life. Today, deep wells are drilled in oil-rich areas to produce natural gas (BP, 2020). The relatively clean nature of natural gas compared to coal and oil, has been a factor in increasing the consumption of this natural resource. Also, by its relationship with other economic sectors and institutions (in the form of inputs or final goods), it plays a significant role in the economic decision-making process and advancing the development goals of countries. According to the British Petroleum (BP) statistical survey of energy consumption globally, the growth of natural gas consumption in 2019 was 2% and amounted to 141.45 exa joul, including a 24.2% share of the energy basket (BP, 2020).

Iran is one of the countries with high natural gas consumption in terms of environmental conditions. In the basket of energy consumed by Iran, natural gas is ranked first among the five primary energy sources (Figure 1). Mainly produced gas is consumed domestically; Iran is the world's fourth -largest gas consumer after the United States, Russia, and China and its gas consumption is 22.49 billion cubic meters (BCM) per day (BP, 2020). The largest gas consumers in Iran are the residential and commercial sectors, which account for 35%, followed by the industrial sector with 27% (EIA, 2020).

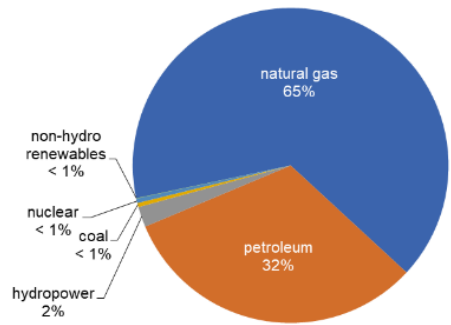

Figure 1: The share of various fuels in Iran's energy consumption basket, 2019 (EIA, 2020)

Natural gas is also used to generate electricity, so that it has a 73% share of electricity production, followed by oil with a 15% share (Figure 2) (EIA, 2020).

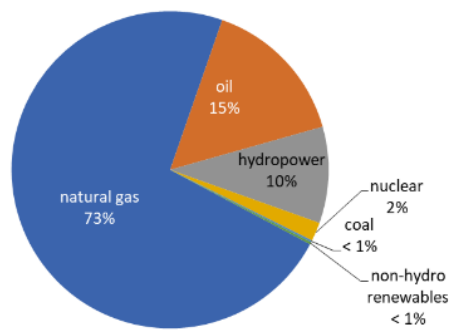

Figure 2: The share of various fuels in Iran's electricity production, 2019 (EIA, 2020)

With the increase in population and energy consumption, consumption patterns have become more critical. In most parts of the world, natural gas consumption is seasonal. Iran is no exception, especially since mountainous areas consume more gas in the cold seasons. It faces problems such as pressure drops and natural gas outages yearly in cold seasons. One of the reasons for the mentioned problems is the consumption pattern of home and industrial subscribers.

Since natural gas is a very suitable substitute for the main products obtained from crude oil refining, including kerosene and liquid gas in the household sector ,therefore, the maximum effort of the country's energy sector is to replace natural gas with petroleum products. The household sector has been one of the essential natural gas-consuming sectors in recent years. In this sector, natural gas consumption consists of two parts: heating and basic consumption. The basic consumption includes household cooking and spa consumption, and this type of consumption has an almost uniform trend throughout the year. There is no significant difference in the amount of consumption in this part based on the change of seasons. Heating consumption relates to the decrease or increase in ambient temperature and increases with the temperature decrease during the year's cold months. This part of household consumption in various regions of the country is different according to the weather conditions of the country. Usually, the mentioned consumption period lasts from two months in warm provinces to six months in cold areas of the country; therefore, heating is a significant part of household consumption.

The above mentioned show that controlling natural gas consumption, especially in the residential sector, which has the largest share in Iran, is a priority. Modification of gas consumption pattern means institutionalizing the correct gas consumption, enhancing living standards, reduces costs in the natural gas sector, and providing a basis for storing this vital energy for future generations. To achieve this, it is necessary predict the customers' consumption to identify the gas consumption pattern and then determine the subscribers' interests in gas consumption (Sangi et al., 2019). Predicting gas consumption is always a significant challenge for a gas company. After that, it will be easier to present programs and solutions to improve the consumption pattern. Data mining and machine learning have different tools for customer segmentation. Therefore, we use these tools to predict consumption, make decisions, and adopt appropriate strategies to deal with customers. The study is conducted in Zanjan province of Iran which due to the cold and dry climate, needs heating in most months of the year, especially in the cold seasons.

The novelty of this paper lies in its ability to train a hybrid BiLSTM-XGBoost model on limited residential natural gas consumption data, a significant deviation from previous works that typically rely on large datasets for model training. While most existing studies in energy forecasting emphasize the availability of vast amounts of data to generate accurate predictions, this study demonstrates that a robust and reliable forecasting model can be developed even with smaller and aggregated datasets. Key to this approach is the efficient use of feature engineering, temporal learning, and external meteorological inputs to supplement the limited consumption data, addressing challenges in data sparsity without compromising predictive performance. Thus, the hybrid model presented here opens new avenues for energy forecasting in contexts where data availability is constrained, offering practical solutions for regions with less extensive or high-resolution energy data.

## 2. Literature Review

Today, energy systems are increasingly using digital information and communication technologies (ICT) to produce, collect, and store large amounts of energy consumption data, therefore, providing an opportunity to implement and analyze big data. Decision-making based on advanced data analysis is essential in forming, operating, and managing intelligent energy systems (Zhou et al., 2017).

Numerous studies have been conducted on predicting energy consumption using various methods and data lengths, including daily, weekly, and monthly time series forecasting. A significant amount of literature is available on this topic (Elshendy et al., 2017). Some researchers have utilized conventional statistical and econometric methods for predicting global crude oil and natural gas consumption, such as the autoregressive integrated moving average (ARIMA) model. Also, traditional techniques such as Generalized Autoregressive Conditional Heteroskedasticity (GARCH) (Ahmed & Shabri, 2014; Wei et al., 2019), Random Walk (RW) (Yu et al., 2017), Error Correction Model (ECM) (Brigida, 2014) are widely used for forecasting purposes in the field (Xiang & Zhuang, 2013; C. lan Zhao & Wang, 2014), and Vector Autoregressive (VAR) models (Ramyar & Kianfar, 2019).

The assumption of a linear relationship between independent variables and dependent variables is the foundation upon which econometric models, such as the ARIMA model, are built. (Cardoso & Cruz, 2016; Kane et al., 2014), have faced criticism for their inability to comprehend the intricacy and non-linear patterns inherent in financial time series data. This lack of understanding has resulted in forecast errors (Sun et al., 2018). Therefore, in order to overcome these difficulties, more intricate and nonlinear artificial intelligence techniques have been implemented recently. These methods encompass Artificial Neural Networks (ANNs) (e.g. Abdullah & Zeng, 2010; Haidar et al., 2008; Ramyar & Kianfar, 2019; Rast, 2001 ); deep learning method (Anagnostis et al., 2020; Bristone et al., 2020; Xue et al., 2019; Zhao et al., 2017); and Support Vector Machine (SVM) (Ahmed & Shabri, 2014; Chiroma et al., 2014; Xie et al., 2006; Yu et al., 2017).

Although integrated forecasting models may not always surpass the best individual forecasting models, the majority of these models demonstrate superior accuracy compared to the average or weakest single forecasting models. This is advantageous as it significantly reduces the likelihood of inaccurate prediction in different practical situations (Wong et al., 2007). Several studies are included below.

Beyca et al., 2019, utilized three commonly used machine learning techniques, namely multiple linear regression (MLR), artificial neural network (ANN), and support vector regression (SVR), to forecast natural gas consumption in Istanbul, the largest city in Turkey in terms of natural gas consumption. The findings revealed that the SVR method outperformed the ANN approach, demonstrating higher reliability and accuracy. The SVR model yielded lower prediction errors, indicating its superiority in time series forecasting of natural gas consumption.

Sangi et al., 2019, in order to discover the consumption pattern, number of time series were extracted from data related to each subscriber for three years. By using data mining method, they concluded that determining the consumption pattern of different uses can play an essential role in reducing costs and help managers to make informed decisions. The contents of the present study seek practical factors and predictions in the natural gas consumption by household subscribers.

Priesmann et al., 2021, review and analyze energy scenarios for future energy devices by combining and separating two sets of data related to weather time series, census data, mobility data, and employment statistics, they used the nomenclature of territorial units for statistics (NUTS2) method as an interface to validate the hybrid method. Finally, they concluded that Jericho-E-based applications' data could interest researchers analyzing energy scenarios in which renewable energy is primarily a substitute for fossil fuels.

Li et al., 2021, utilized intelligent cluster analysis and reviewed the consumption bills of 3,995 households in Hefei province, China, to investigate domestic gas consumption patterns. The findings revealed that these patterns could be categorized into four types: single-point spike, double-point flat-peak, micro-peak, and linear. Single-point spike type customers and double-point flat-peak type consumers are classified as high gas consumption consumers at the second and third levels of the consumer hierarchy. A significant number of houses are comprised of consumers who utilize wall-hung boilers.

Du et al., 2022, conducted a study to tackle the limitations of traditional time series prediction in forecasting natural gas consumption. They investigated various regions in China and presented a new approach, comparing it to advanced methods like long short-term memory (LSTM) and convolutional neural network long short-term memory (CNN-LSTM). The findings demonstrated that the proposed method, when combined with other modules, substantially enhanced the accuracy and reliability of predictions. It achieved a notable reduction in the mean absolute percentage error, ranging from 0.235% to 10.303%, outperforming alternative models.

Singh et al., 2023, conducted an analysis of the annual natural gas consumption in the United States from 1980 to 2020. They employed three modeling techniques, namely multiple linear regression (MLR), artificial neural network (ANN), and support vector machine (SVM), to predict the consumption. The results indicated that the ANN model achieved the best performance, with a mean absolute percentage error (MAPE) of 0.022 for the training dataset and 0.03 for the testing dataset.

## 3. Artificial neural network and machine learning

The artificial neural network was first proposed by Warren McCulloch and Walter Pitts around 1943 and in the final years of World War II. Artificial neural networks (ANNs) are frequently utilized as a standalone model or as part of a hybrid forecasting model. Although artificial neural networks are designed based on the biological structure of neurons, they cannot replace real neurons; however, they have the power to solve various problems.

The concept of artificial neural network in machine learning is derived from artificial intelligence (AI), which can simulate the human brain system along with its neural network and create a system that can learn, make mistakes, and learn from its mistakes. Recurrent Neural Networks (RNNs) are a potent and durable type of NN that features an internal memory to remember crucial information about previous inputs. This is necessary in order to make precise predictions regarding what will occur in the future. RNNs have a similar structure to other NN types, with modifications that account for hidden layers from previous time steps when processing the current time step. With the exclusion of Recurrent Neural Networks (RNNs), the majority of neural networks (NNs) lack a memory to handle individual inputs presented to them, and there is no continuation of a previous state between inputs (Chollet & Allaire, 2018). A neural network, often known as a NN, is made up by three layers that are input, hidden, and output layers. Recurrent Neural Networks (RNNs) are a robust and resilient form of neural networks that possess an internal memory to retain important information from past inputs. It is crucial to deliver precise predictions regarding future events. Recurrent Neural Networks (RNNs) have a structure that is akin to different types of Neural Networks. However, RNNs incorporate adjustments to accommodate hidden layers from past time steps while processing the current time step. There are two versions of RNNs: Long Short-Term Memory (LSTM) and Gated Recurrent Unit (GRU). The transmission of information within these networks (such as RNN and its variations) is capable of moving in any direction between and within groups. A significant obstacle encountered with simple RNNs is the need to preserve information from inputs that have detected multiple time steps in the past. However, the issue of vanishing gradient poses a hurdle in learning these long-term dependencies. The LSTM and GRU models were developed to tackle this problem. These models are extensively utilized because of their superiority over alternative methods and their ability to precisely forecast periods of peak flow (Apaydin et al., 2020; Busari & Lim, 2021).

### 3.1. Long-short term memory (LSTM)

The LSTM model that developed by Hochreiter & Schmidhuber, 1997, is one of the artificial neural network models created due to problems that RNN had in the vanishing gradient. This model includes three gates: input, forget, and output. The studies show this model had brought a much better performance and a much higher speed in the learning process. LSTM network has several essential parts (3), which are: 1) hidden part, 2) input part, 3) internal part, 4) input gate, 5) forget gate, 6) output gate. The structure of LSTM allows for solving several problems related to sequence models.

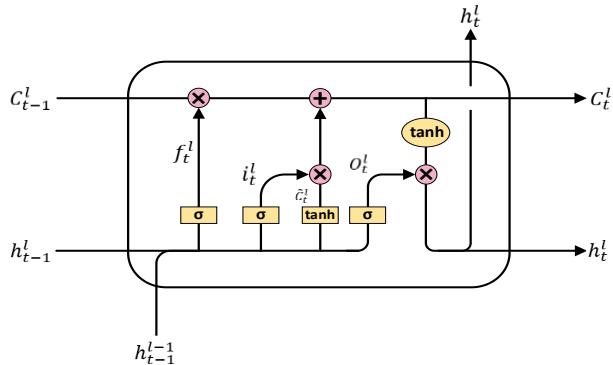

Figure 3: LSTM cell structure (Hochreiter & Schmidhuber, 1997)

### 3.2 Bidirectional LSTM

Bidirectional Long Short-Term Memory (BiLSTM) networks have emerged as a powerful variant of recurrent neural networks, offering enhanced capabilities in capturing context from both past and future states (Graves & Schmidhuber, 2005). Unlike traditional LSTMs that process sequences in a forward direction, BiLSTMs utilize two separate hidden layers to process input sequences in both forward and backward directions (Schuster & Paliwal, 1997). This bidirectional approach allows the network to leverage both preceding and succeeding context, making BiLSTMs particularly effective for tasks such as named entity recognition, part-of-speech tagging (Wang & Lin, 2014), and machine translation (Cho et al., 2014). The architecture of a BiLSTM typically consists of an input layer, forward and backward LSTM layers, and an output layer that combines information from both directions.

According to Figure 4 description of the architecture of model is as follows:

1) Input Layer: Now labeled with ($x_1$, $x_2$, ..., $x_t$) to represent the input sequence.
2) Embedding Layer: Added between the input and LSTM layers to show how input is typically transformed into dense vectors.
3) LSTM Cells: Both forward and backward LSTM layers now show a detailed view of an LSTM cell, including:

    - Forget gate (f gate)
    - Input gate (i gate)
    - Output gate (o gate)
    - tanh layer for creating new cell state candidates

4) Directional Flow: Clearer representation of forward and backward processing.
5) Concatenation Layer: Explicitly shows how forward and backward hidden states are combined.
6) Output Layer: Includes the SoftMax activation function often used for classification tasks.

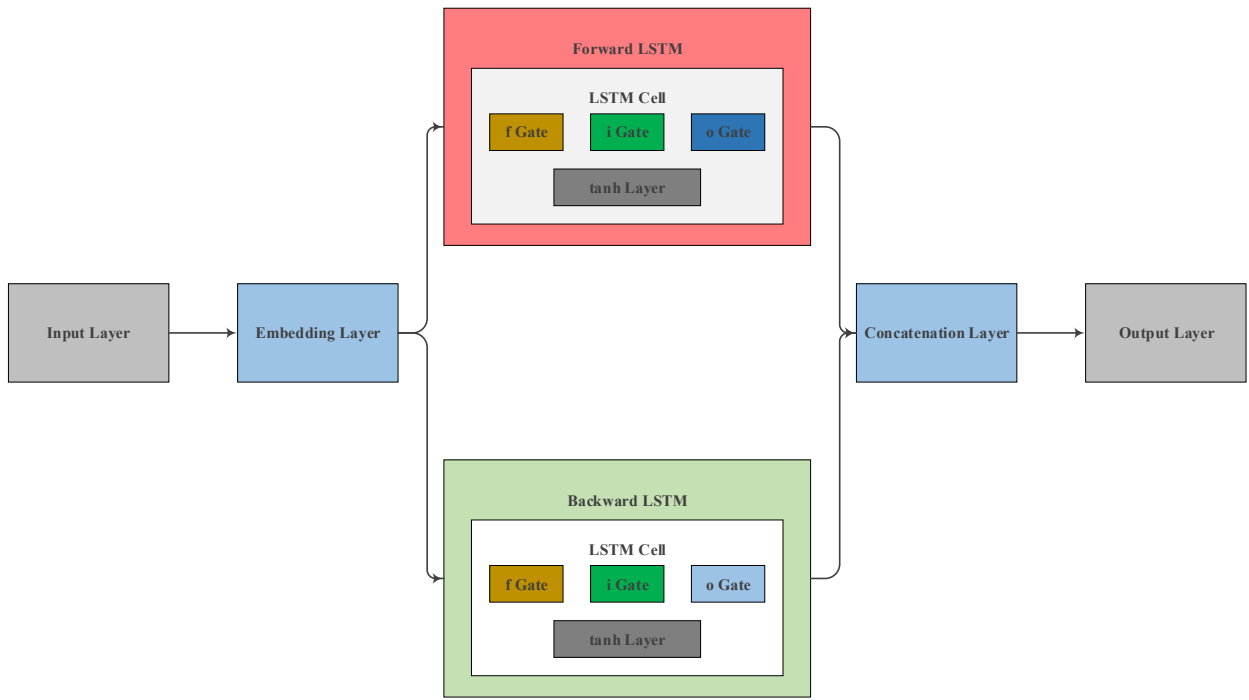

Figure 4: BiLSTM cell structure (Chung et al., 2014)

## 3.3 Gated recurrent unit (GRU)

Gated recurrent units (GRUs) are a gating mechanism in recurrent neural networks, introduced in 2014 by Chung et al. GRU layers are almost similar to LSTM layers, except that the input and forget gates are combined. Also, cell state and hidden state are merged. Therefore, GRU can be considered a model with fewer parameters than LSTM. This feature can run the model faster than before. This model includes two gates consisting of reset and update gates (Figure 5). The reset gate receives the input data, combines it with previous data, and decides which data should be forgotten. Next, the update gate determines the amount of last memory to store data. The update gate works similar to the forget gate and the input gate in the LSTM model. There are differences between LSTM and GRU models:

1) LSTM model has three gates which include forget, input, and output gates, while the GRU model has two reset and update gates.
2) The GRU has an internal memory that is equivalent to the revealed hidden state.
3) In contrast to LSTM, GRU is able to reveal all of the memory and hidden layers.
4) GRU performs better on smaller data while LSTM performs better on larger data.

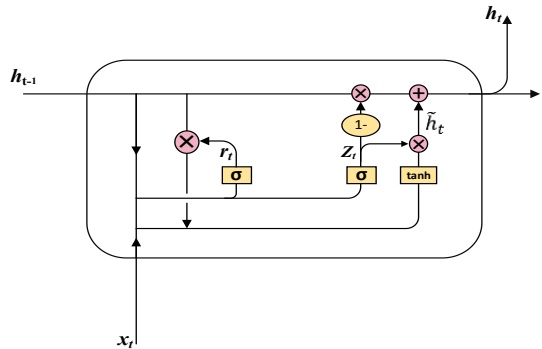

Figure 5: GRU cell structure (Chung et al., 2014)

## 3.4 Extreme gradient boosting (XGBoost)

XGBoost (Extreme Gradient Boosting) is a highly efficient and scalable implementation of gradient tree boosting, a widely used machine learning algorithm known for its predictive accuracy in various tasks. It provides state-of-the-art performance in many machine learning challenges, including classification, regression, and ranking problems(Chen & Guestrin, 2016) .

One of the key strengths of XGBoost is its ability to handle sparse data efficiently. It employs a sparsity-aware algorithm that makes it faster in situations with missing values or zero entries, a common occurrence in many real-world datasets. This feature is particularly useful in high-dimensional datasets where most values are zero, such as in one-hot encoded data (Chen & Guestrin, 2016; Panda et al., 2009; Tyree et al., 2011; Ye et al., 2009).

Additionally, XGBoost introduces a novel weighted quantile sketch algorithm, which allows it to handle weighted data more effectively during the tree-building process. This method ensures that the model can compute more accurate splits when learning from large-scale, imbalanced datasets (Chen & Guestrin, 2016; Panda et al., 2009; Tyree et al., 2011; Ye et al., 2009).

The model is optimized for performance in distributed settings, with support for parallel computation across multiple machines. It includes out-of-core computing capabilities, enabling the processing of datasets that are too large to fit in memory by efficiently managing disk usage and computational resources (Chen et al., 2013, 2015).

## 4. Methodology

The methodology employed in this study is outlined in Figure 6, illustrating the various stages, including data collection, pre-processing, feature engineering, splitting the dataset into training and testing sets, learning process, and evaluation of the models, and extracting statistical results to assess predictive performance.

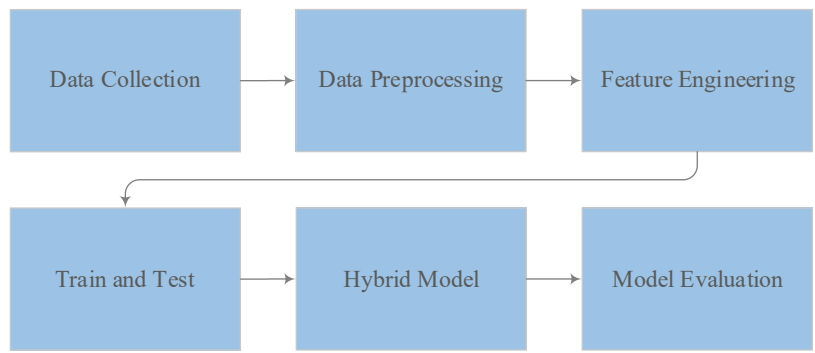

Figure 6: Flowchart of the methodology

### 4.1. Data collection

In conducting this research, the natural gas subscribers of Zanjan province have been selected as the case study. Zanjan is a historical town situated along the historic Silk Road. Zanjan, situated in the mountainous region of Iran, experiences brief, temperate summers and severe, lengthy winters. As a result, it has become renowned as one of the premier places for winter tourism. While agriculture plays a significant role in the city's economic activity, it is the industrial sector that dominates economic production. From this province, 5 main cities have chosen for data collection (Statistical Center of Iran, 2022). The names of the cities chosen are as follows:

1. Abbar
2. Abhar
3. Khoramdare
4. Mahneshan
5. Zanjan

The selected period for gas consumption is six years (2017- 2022) and on a monthly basis, which has been collected from The Statistical Center of Iran (SCI) (Statistical Center of Iran, 2022).
In the next, weather information for the mentioned period of time have collected from National Weather Organization. Features that we chose from weather database are: Mean Relative Humidity, Max Relative Humidity, Min Relative Humidity, Mean Temperature, Max

Temperature, Min Temperature, Max Absolute Temperature, Min Absolute Temperature, Rainfall (mm), Freezing.

### 4.2. Data pre-processing

In this study, thorough data pre-processing was essential to ensure the robustness of the model. We address two main challenges: handling missing data and normalizing the dataset for improved model performance.

### 4.2.1. Handling Missing Data

Our dataset contained missing values, particularly from some cities where data was not consistently recorded. Initially, we considered two approaches: either removing cities with significant gaps in their data or filling in the missing values using statistical methods like mean, median, or constant values (such as zeros).

However, for more accuracy and continuity, we opted for a more advanced strategy. Time-series data, in particular, benefits from interpolation methods, which can predict missing values based on the trends of adjacent data points. We employed time-series interpolation to estimate missing values, ensuring that the temporal relationships in the data remained intact. This method better preserves the structure of the dataset compared to simple imputation methods like mean or median, especially in cases where the missing data is sequential rather than randomly distributed.

In cases where interpolation was not feasible or data was completely missing for certain regions, we considered advanced imputation techniques like using machine learning models to predict the missing values. By using more sophisticated methods, we aimed to maintain the overall integrity of the dataset and avoid introducing bias or inaccuracies.

### 4.2.2. Data Scaling

Since our models (LSTM and GRU) are sensitive to the scale of input data, it was important to normalize the features. Initially, we applied MinMaxScaler from the SKLearn library, which scales data to a range between 0 and 1. This method works effectively for ensuring that all features contribute equally to the model's performance.

However, depending on the distribution of the data, MinMax scaling can sometimes compress the variability of the data if outliers are present. To account for this, we evaluated other normalization techniques, such as Z-score normalization (also known as standardization). Z-score normalization transforms the data so that it has a mean of 0 and a standard deviation of 1, which can be particularly useful when the data contains outliers, as it reduces the influence of extreme values on the model.

Additionally, for variables that exhibited right-skewed distributions, we applied log transformations. This step helped stabilize the variance, making the data more normally distributed, which in turn improved the performance of the machine learning models. The log transformation also allowed us to model data with exponential growth patterns more effectively.

By combining interpolation for missing data with appropriate normalization techniques, we ensured that the dataset was both complete and well-scaled, allowing the models to learn more efficiently from the data.

### 4.3 Feature Engineering

Feature engineering plays a crucial role in enhancing the model's predictive capabilities. The process begins with the original dataset, which includes basic features such as energy usage, relative humidity, temperature, and precipitation, it also contains limited records which need following process to prepare for training the model. To capture the temporal aspects of energy consumption patterns, we derive several time-based features from the date column:

- Month: Extracted to capture seasonal patterns in energy usage that may occur on a monthly basis.
- Day: Represents the day of the month, which may reveal patterns related to billing cycles or monthly routines.
- Day Of Week: Encodes the day of the week (0-6), allowing the model to learn weekly patterns in energy consumption.
- Quarter: Represents the quarter of the year, potentially capturing broader seasonal trends.

These engineered features allow the model to learn from cyclical patterns in energy usage that may be tied to daily, weekly, monthly, or seasonal factors. By including these temporal features, we provide the model with important context about the time-dependent nature of energy consumption. In addition to feature extraction, we also apply a logarithmic transformation to the target variable (Usage). This transformation helps to stabilize the variance in the data and can improve the model's performance, especially when dealing with energy consumption data that often exhibits right-skewed distributions.

The feature engineering process is followed by normalization using MinMaxScaler, which scales all features to a range between 0 and 1. This step is crucial for ensuring that all features contribute equally to the model and for optimizing the performance of both the LSTM and XGBoost components.

By combining domain knowledge with data-driven feature engineering, we aim to provide our hybrid model with a rich, informative feature set that captures both the direct measurements and the underlying temporal patterns in energy consumption.

### 4.4. Training and testing

The training and testing of the models were conducted using a rigorous process to ensure accurate and reliable performance evaluations. The dataset, which consisted of household gas consumption data for five cities, was divided into training and testing subsets to validate the predictive capabilities of each model. The training set was used to train the models, including LSTM, GRU, and the hybrid BiLSTM-XGBoost, while the testing set was used to evaluate their performance on unseen data.

We utilized Python for model implementation, with the Keras library for deep learning models and Scikit-Learn for data pre-processing. The LSTM, GRU, and BiLSTM-XGBoost models were configured with hyperparameters optimized through grid search to achieve the best possible results.

The models were trained using a high-performance machine with the following specifications: an Intel Core i9 processor, 32GB of RAM, and an NVIDIA RTX 3090 GPU. The LSTM and GRU models were trained for 500 epochs with a batch size of 32, while the hybrid BiLSTM-XGBoost model utilized a combination of recurrent layers and gradient boosting, with an emphasis on capturing both temporal dependencies and feature interactions.

The training and testing process highlighted the strengths and limitations of each model, with the hybrid approach providing the best overall performance, particularly in cases with limited data availability.

### 4.5 Hybrid XGBoost-BiLSTM

This study presents a novel hybrid approach for energy usage prediction, combining the strengths of Long Short-Term Memory (LSTM) networks and eXtreme Gradient Boosting (XGBoost). The LSTM component, a type of recurrent neural network, excels at capturing temporal dependencies in time series data, while XGBoost, an ensemble learning method, is adept at handling non-linear relationships and feature interactions. By integrating these two models, we aim to leverage their complementary strengths to enhance prediction accuracy and robustness.

The hybrid model (Figure 7) architecture begins with data preprocessing, including handling missing values, feature engineering, and normalization. The LSTM component consists of multiple bidirectional LSTM layers with dropout for regularization, allowing the model to learn complex temporal patterns in both forward and backward directions. Concurrently, the XGBoost model is trained on the same feature set, capturing non-linear relationships and feature importances. The final prediction is obtained by averaging the outputs of both models, potentially mitigating individual model weaknesses and producing more stable and accurate forecasts.

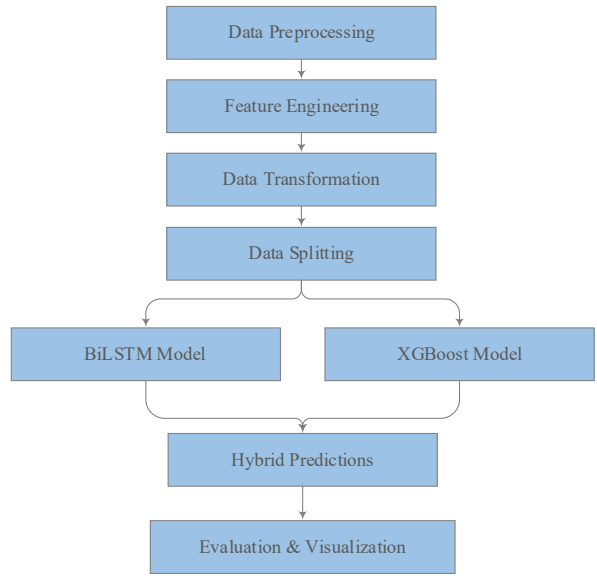

Figure 7: Hybrid BiLSTM-XGBoost structure

### 4.5. Model Evaluation

Multiple methodologies exist for assessing the effectiveness of time-series forecasting models, such as error computations and determining the statistical significance of these errors. To assess the performance of our time-series models, we employed several evaluation metrics to quantify prediction accuracy and error. The key metrics used were the Root Mean Squared Error (RMSE), Mean Absolute Percentage Error (MAPE), and Mean Percentage Error (MPE). These metrics provide a comprehensive understanding of the prediction performance by capturing both the magnitude and the direction of the forecast errors.

Root Mean Squared Error (RMSE):

RMSE is one of the most commonly used metrics in forecasting, as it gives greater weight to larger errors by squaring them. It helps in measuring the overall accuracy of the model by evaluating the standard deviation of the prediction errors. RMSE is particularly sensitive to outliers and larger discrepancies between predicted and actual values.

The formula for RMSE is:

$$RMSE = \sqrt{\frac{1}{n} \sum_{i=1}^{n} (P_i - A_i)^2}$$

Where:
- $A_i$ is the actual value
- $P_i$ is the predicted value
- $n$ is the number of observations

Mean Absolute Percentage Error (MAPE):

MAPE calculates the average absolute percentage difference between actual and predicted values, making it a useful metric for understanding the relative error magnitude. Unlike RMSE, it does not emphasize larger errors and provides a clear percentage measure of accuracy.

The formula for MAPE is:

$$MAPE = \frac{1}{n} \sum_{i=1}^{n} \left| \frac{A_i - P_i}{A_i} \right| \times 100$$

Where:
- $A_i$ is the actual value
- $P_i$ is the predicted value
- $n$ is the number of observations

Mean Percentage Error (MPE):

MPE, unlike MAPE, retains the direction of errors, allowing us to understand whether the model tends to overestimate or underestimate the actual values. A negative MPE indicates systematic overprediction, while a positive MPE suggests underprediction on average.

The formula for MPE is:

$$MPE = \frac{1}{n}\sum_{i=1}^{n}\frac{A_i - P_i}{A_i} \times 100$$

Where:

- $A_i$ is the actual value
- $P_i$ is the predicted value
- $n$ is the number of observations

These metrics are calculated on both training and test sets to assess the model's predictive power and generalization capability. Additionally, we visualize the model's performance through various plots, including actual vs. predicted energy usage and feature importance, providing insights into the model's behavior and the relative impact of different variables on energy consumption predictions.

## 5. Results and Discussion

### 5.1. Comparative Analysis of Forecasting Model Performance

Figures 8,9,10 present the forecasting results for the three models evaluated: GRU, LSTM, and the hybrid BiLSTM-XGBoost model. The results indicate that the hybrid BiLSTM-XGBoost model outperformed the other models in terms of predictive accuracy, as reflected in closer alignment with the perfect prediction line for both training and testing data.

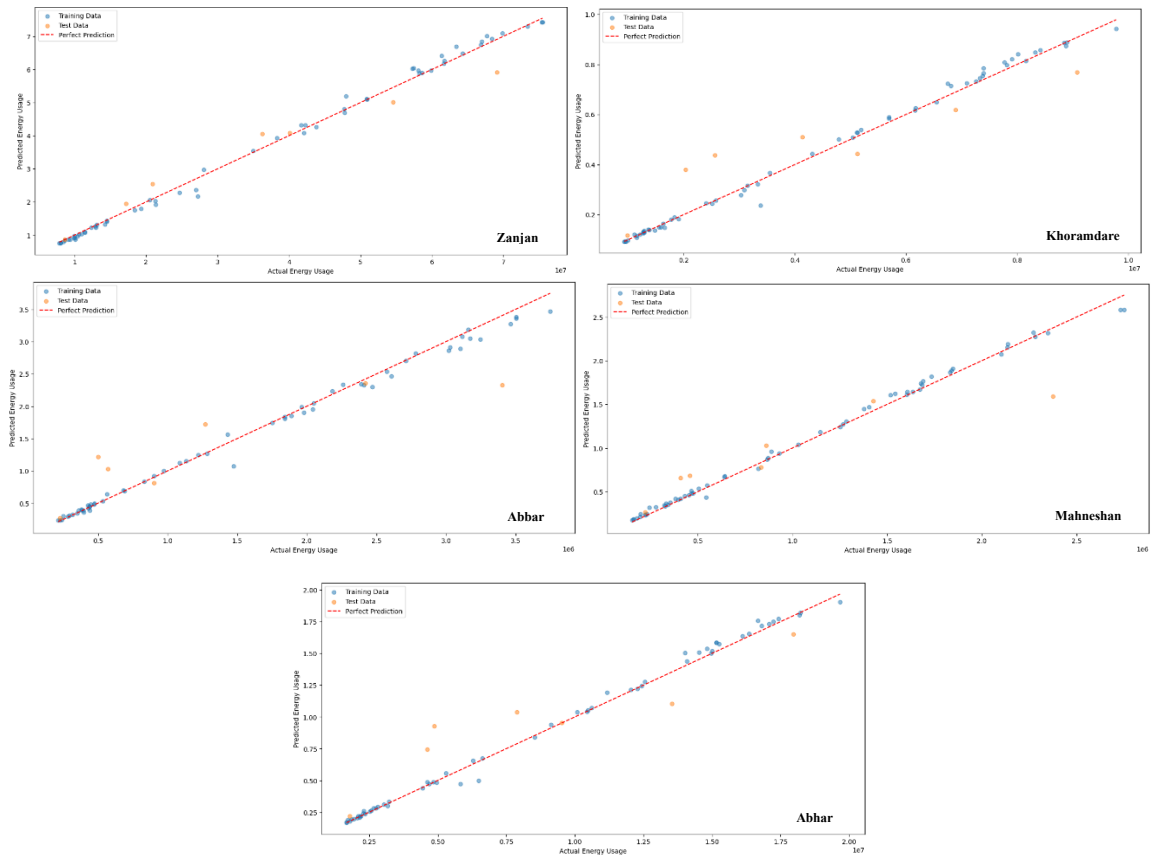

Figure 8: Comparison of actual and forecast values for hybrid model

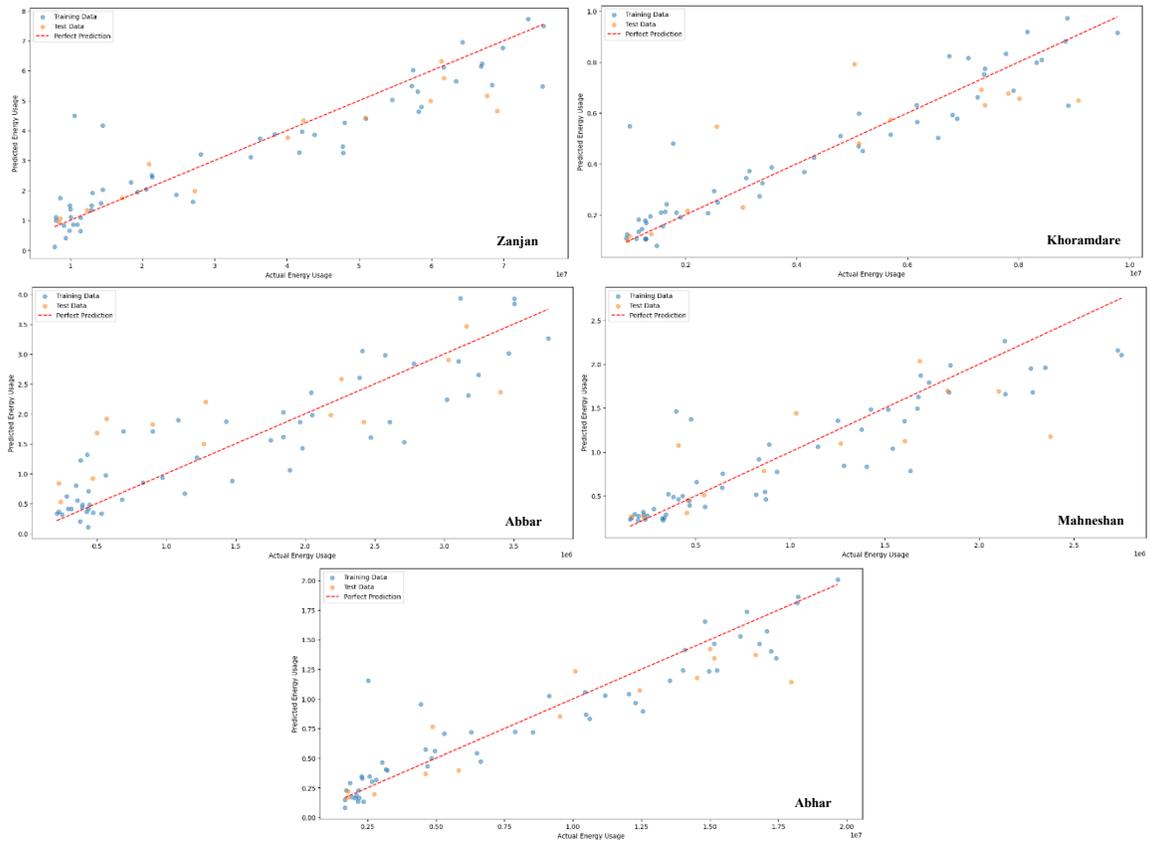

Figure 9: Comparison of actual and forecast values for LSTM

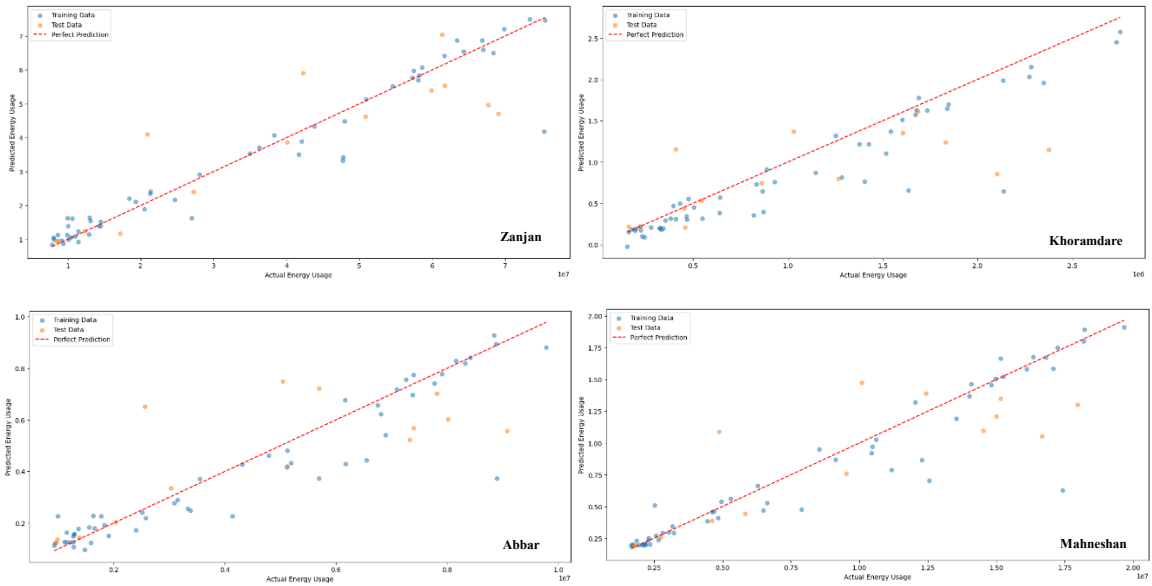

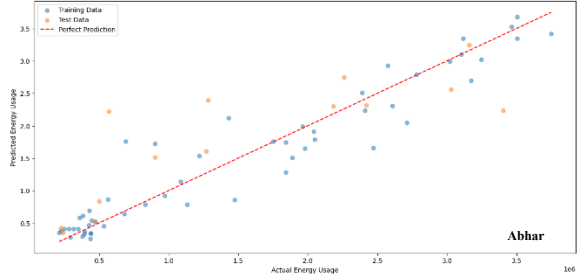

Figure 10: Comparison of actual and forecast values for GRU

The scatter plots comparing actual versus predicted energy usage reveal that the hybrid BiLSTM-XGBoost model consistently provides predictions that align closely with the perfect prediction line. This indicates that the hybrid model effectively captures both temporal dependencies and complex non-linear relationships within the gas consumption data. Its superior performance is evident in the lower dispersion of points from the line, particularly for the test dataset, which highlights its ability to generalize well to unseen data.

The GRU model, while not as accurate as the hybrid model, demonstrated robust performance, with many data points closely aligned with the perfect prediction line. The GRU model's simplicity allows it to effectively handle limited data while still maintaining good predictive accuracy. However, it lacks the feature interaction capabilities present in the hybrid model, which limits its overall performance.

The LSTM model showed the greatest deviation from the perfect prediction line, with a noticeable spread in both training and testing data points. This suggests that LSTM struggles with the complexity of the dataset, potentially due to overfitting or underfitting issues. Its performance was the weakest among the three models, indicating that it may not be the best choice for this type of data, especially when the dataset is limited.

The results demonstrate that the hybrid BiLSTM-XGBoost model is the most suitable for predicting domestic gas consumption patterns, particularly in areas with complex seasonal and geographical variations. These findings underscore the importance of hybrid models in capturing the diverse factors influencing energy consumption and highlight their effectiveness even in scenarios with limited data availability.

**5.2. Hybrid Model Counterintuitive Feature Selection: Challenging the Primacy of "Usage"**

The feature importance charts derived from XGBoost models across five cities reveal a recurring tension between autoregressive dependencies (e.g., historical "Usage") and environmental or temporal drivers. While conventional wisdom might prioritize "Usage" as the dominant predictor in time-series forecasting—given its direct autoregressive relationship with future demand—the hybrid model's feature selection logic introduces a nuanced counterpoint. For instance, in Zanjan, "Freezing Days" emerges as the most critical feature (importance ≈0.40), dwarfing "Usage" (ranked third, importance ≈0.25). Similarly, in Khoramdare, "Usage" ranks first but shares significance with "Min Absolute Temperature" and "Freezing Days," suggesting the hybrid model integrates climatic extremes to temper reliance on lagged usage data. This divergence implies that the hybrid framework actively suppresses overfitting risks by redistributing weight to exogenous variables, even when autoregressive signals appear strong (Figure 11, Table 1).

### 5.2.1. Mechanisms Behind the Hybrid Model's Feature Prioritization

The hybrid model's architecture likely combines gradient-boosted trees (XGBoost) with domain-specific rules or post-hoc feature engineering to counteract biases toward autoregressive features. For example:

- **Environmental Overrides**: In colder climates like Zanjan, extreme weather events ("Freezing Days") may disrupt typical usage patterns, rendering historical "Usage" less predictive. The hybrid model amplifies such features to capture acute demand spikes (e.g., heating surges during cold snaps).
- **Temporal Contextualization**: While "Month" is universally important, its interaction with "Usage" varies. In Abbar, "Usage" and "Max Temperature" share high importance (≈0.6 and ≈0.5, respectively)(Figure 11, Table 1), but their combined effect may be modeled multiplicatively (e.g., higher temperatures exacerbate cooling demand, altering the baseline usage trend). The hybrid system could explicitly encode these interactions, reducing the standalone weight of "Usage."
- **Regularization Effects**: By penalizing over-reliance on "Usage" (a potential leakage risk if misconfigured), the hybrid model ensures robustness. In Mahneshan, "Min Temperature" and humidity metrics overshadow "Usage," indicating that the model prioritizes physical drivers over historical trends when they offer comparable predictive power (Figure 11, Table 1).

This model integrates time-series learning and feature importance analysis, yielded fewer errors without explicitly incorporating altitude data. However, the model's predictions remain consistent with the altitude table, as higher gas consumption is observed in regions at higher elevations (e.g., Zanjan) compared to lower-altitude urban centers like Abbar, underscoring that climatic and geographic factors are effectively captured through other variables (Table 1).

### 5.2.2. Implications for Model Interpretability and Deployment

The hybrid model's rejection of "Usage" as the omnipotent feature challenges traditional forecasting paradigms. In Abhar, for instance, "Rain (mm)" ranks fourth (importance ≈0.3), surpassing "Usage" (importance ≈0.35) only marginally—a subtle but critical recalibration. This suggests the model incorporates real-time environmental feedback loops (e.g., rainfall reducing outdoor water usage) that historical data alone cannot encapsulate. For policymakers, this underscores the need to pair infrastructure planning with climate adaptation strategies, as static usage histories may fail under shifting environmental conditions (Figure 11, Table 1).

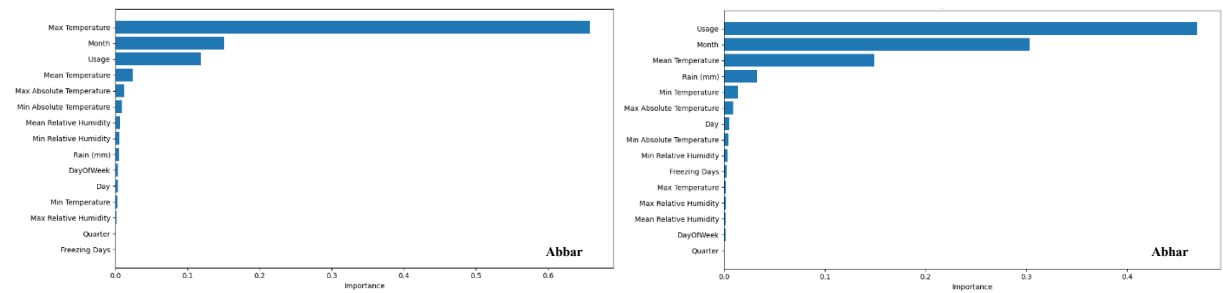

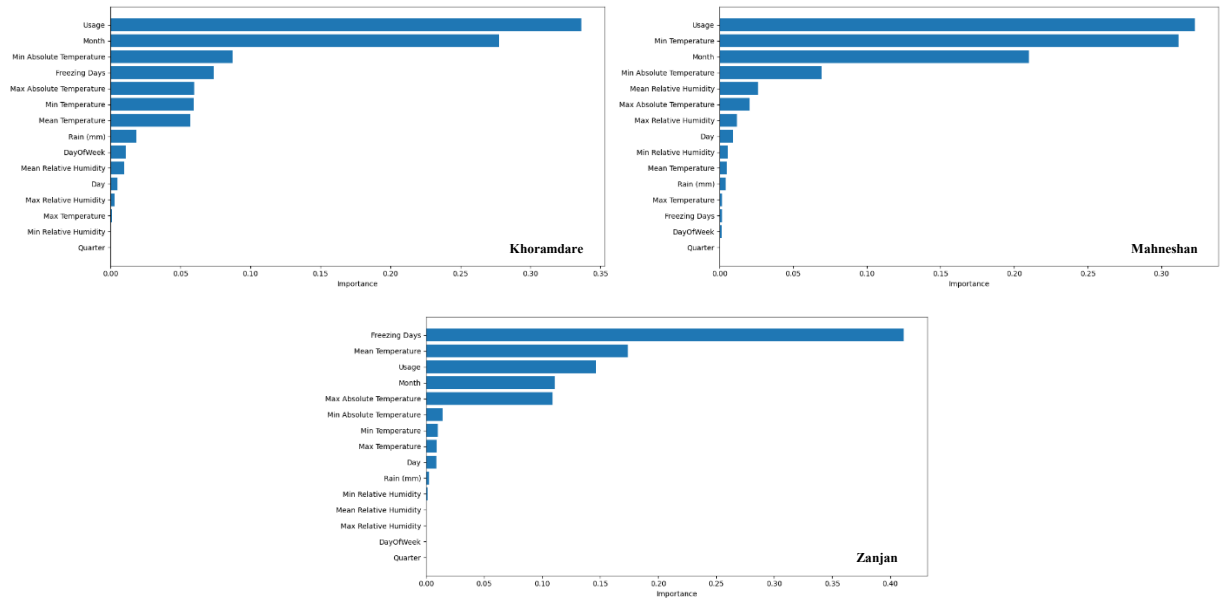

Figure 11: Hybrid model feature selection ranking charts based on each city

| City | Elevation (meters) |
|---|---|
| Abbar | 1,336 |
| Abhar | 1,573 |
| Khoramdare | 1,655 |
| Mahneshan | 1,300 |
| Zanjan | 1,642 |

Table 1: Cities elevation

The hybrid model's feature selection logic disrupts the assumption that "Usage" is invariably the foremost predictor. By dynamically weighting features based on localized climatic stressors, temporal granularity, and regularization constraints, it achieves a balance between autoregressive fidelity and environmental responsiveness (Table 1). This approach not only enhances generalizability across diverse geographies but also aligns with sustainable resource management principles, where understanding why demand fluctuates (e.g., freezing days, humidity) proves as vital as tracking how much it fluctuates. Future work could explore hybrid architectures that further disentangle these drivers, ensuring models remain interpretable while embracing complexity.

**5.3. Quantitative Performance Evaluation: RMSE Benchmarking and Error Analysis**

From the evaluation results (Table 2, 3), it is evident that the BiLSTM-XGBoost model showed the best overall performance in terms of RMSE, with consistently lower values across most cities. For example, in Abhar, the BiLSTM-XGBoost model achieved an RMSE of 1,500,470, which was significantly lower compared to the LSTM (3,311,754) and GRU (3,484,798) models. This trend was consistent across several other locations, indicating the strength of the hybrid model in capturing complex non-linear patterns in the gas consumption data.

In terms of MAPE, the BiLSTM-XGBoost model also performed better, with lower error rates indicating higher predictive accuracy. For instance, in Khoramdare, the MAPE for BiLSTM-XGBoost was 8.42%, compared to 20.27% for LSTM and 33.33% for GRU. This suggests that the hybrid approach was more effective at generalizing across the dataset, likely due to its ability to capture both temporal dependencies (through BiLSTM) and non-linear relationships (through XGBoost).

The GRU model, while not as accurate as BiLSTM-XGBoost in terms of RMSE and MAPE, demonstrated more consistent performance across different cities, with fewer instances of significant overprediction or underprediction, as reflected in the MPE values. For example, in Abbar, the GRU model had an MPE of -46.27%, which indicates a lower bias compared to LSTM (-57.31%). This suggests that GRU may be more robust in scenarios with smaller datasets or limited features. However, contrary to previous research that suggested hybrid models are unsuitable for limited datasets, the BiLSTM-XGBoost model in this study outperformed GRU even with limited data, indicating its compatibility and effectiveness in handling smaller datasets due to its ability to capture complex temporal and non-linear relationships.

|      | Cities     | BiLSTM-XGBoost |        |        | LSTM      |        |         | GRU      |        |         |
|------|------------|----------------|--------|--------|-----------|--------|---------|----------|--------|---------|
|      |            | RMSE           | MAPE   | MPE    | RMSE      | MAPE   | MPE     | RMSE     | MAPE   | MPE     |
| Test | Abbar      | 343917.6       | 16.43% | 9.12%  | 644213.04 | 70.10% | -57.31% | 853031.3 | 58.78% | -46.27% |
|      | Abhar      | 1500470        | 9.63%  | 1.15%  | 3311754   | 27.01% | 6.82%   | 3484798  | 31.74% | -11.51% |
|      | Khoramdare | 708504.6       | 8.42%  | 5.38%  | 1191645   | 20.27% | -3.73%  | 1720774  | 33.33% | -11.22% |
|      | Mahneshan  | 327506.4       | 21.28% | 13.67% | 419467.4  | 38.72% | -23.60% | 505718.3 | 40.47% | -20.85% |
|      | Zanjan     | 6908990        | 10.56% | -1.78% | 9208265   | 15.39% | 4.22%   | 11904809 | 25.89% | 7.71%   |

Table 2: RMSE, MAPE, and MPE values for test section one step ahead forecasts

|       | Cities     | BiLSTM-XGBoost |       |        | LSTM     |        |         | GRU      |        |        |
|-------|------------|----------------|-------|--------|----------|--------|---------|----------|--------|--------|
|       |            | RMSE           | MAPE  | MPE    | RMSE     | MAPE   | MPE     | RMSE     | MAPE   | MPE    |
| Train | Abbar      | 84749.86       | 4.57% | -2.36% | 438212.1 | 34.47% | -12.35% | 313752.6 | 22.07% | 0.85%  |
|       | Abhar      | 781286.5       | 3.25% | 0.63%  | 1981584  | 19.29% | -1.12%  | 1981174  | 14.53% | -2.31% |
|       | Khoramdare | 485352.4       | 5.44% | -0.37% | 1065973  | 32.06% | -20.34% | 957813.6 | 20.87% | -8.15% |
|       | Mahneshan  | 56668.69       | 4.19% | -0.59% | 338925   | 36.80% | -12.03% | 267886.4 | 22.59% | -9.49% |
|       | Zanjan     | 986544.4       | 2.61% | -0.18% | 8209389  | 27.24% | -9.95%  | 6777196  | 16.22% | 6.29%  |

Table 3: RMSE, MAPE, and MPE values for for train section one step ahead forecasts

## 6. Generalizability and Adaptation to Diverse Contexts

The proposed hybrid BiLSTM-XGBoost framework is designed with flexibility in mind, making it well suited for application beyond the current case study of residential natural gas consumption in Zanjan province. The methodological steps—ranging from data collection and pre-processing to feature engineering and model training—are inherently modular and can be tailored to suit diverse datasets and regional contexts.

## 6.1. Adaptation to Different Geographical Regions

While this study focuses on five cities within a single province in Iran, the underlying modeling approach is largely independent of geographic specificity. Researchers can extend the framework by incorporating region-specific variables that affect energy consumption. For example, in other countries or regions, additional factors such as local socioeconomic indicators, building insulation standards, urban density, or even different climatic variables (e.g., humidity levels, wind speed) may be influential. By recalibrating the feature engineering process to include these localized parameters, the model can capture the unique energy consumption dynamics of different geographical areas.

Furthermore, cross-regional comparative studies can be conducted by training the model on multi-regional datasets. This would not only validate the robustness of the hybrid approach but also facilitate transfer learning, where knowledge acquired from data-rich regions can be leveraged to improve predictions in areas with sparser data.

## 6.2. Flexibility in Data Resolution

The data pre-processing techniques employed in this study—such as time-series interpolation, normalization, and temporal feature extraction—are independent of the data's temporal granularity. Although our analysis is based on monthly aggregated data, the framework can be readily adapted for higher-resolution datasets (e.g., daily or hourly measurements). In contexts where more granular data are available, the model can be fine-tuned to capture short-term fluctuations and transient events that might be obscured in coarser datasets.

Conversely, in regions where only aggregated or irregular data are accessible, the hybrid model's ability to handle limited datasets through advanced interpolation and robust normalization techniques ensures that meaningful predictive insights can still be derived. This adaptability broadens the applicability of the model to various energy systems with differing data availability and quality.

## 6.3. Future Directions for Enhancing Generalizability

To further enhance the model's generalizability, future research should focus on:

- **Transfer Learning:** Leveraging pre-trained models from regions with abundant data to improve predictions in data-sparse environments.
- **Cross-Validation Across Regions:** Conducting empirical studies across diverse geographical and climatic settings to assess model robustness.
- **Incorporation of Additional Variables:** Integrating socio-economic and infrastructural indicators that are relevant in different contexts to refine predictive performance.
- **Dynamic Feature Engineering:** Developing adaptive feature selection methods that can automatically adjust to the specific characteristics of regional datasets.

By addressing these areas, the hybrid BiLSTM-XGBoost model can be positioned as a versatile tool for global energy forecasting, ultimately aiding policymakers and energy managers in designing region-specific strategies for sustainable resource management.

## 7. Conclusion and Policy Implications

Iran is one of the world's most influential countries in energy, ranking as the third-largest natural gas producer. However, the rapid increase in residential natural gas consumption is straining the nation's resources, making it essential to study and predict customer usage patterns to develop effective strategies for reducing consumption. To address these challenges, we applied advanced data analysis and machine learning tools. These methods enable a more detailed examination of consumption behaviors compared to traditional approaches, offering valuable insights that can help policymakers design targeted interventions for efficient and sustainable energy management. This study analyzed household gas consumption across five cities in Zanjan province over a six-year period (2017–2022) using advanced time-series analysis with LSTM, GRU, and a hybrid model. Notably, the hybrid model, which integrated temporal learning with feature importance analysis, yielded fewer errors, underscoring that climatic and geographic factors (e.g., high-altitude areas like Zanjan consuming significantly more gas than urban centers like Abhar) are key drivers of consumption, beyond what population density alone can explain.

A primary challenge was the reliance on monthly aggregated gas data, which masked short-term fluctuations—such as daily or weekly spikes driven by abrupt weather changes. To mitigate this limitation, high-resolution meteorological variables (daily temperature extremes, freezing days) and engineered features (e.g., "average weekly temperature") were incorporated to approximate finer temporal drivers. However, this approach introduced dependencies on external weather datasets that sometimes suffer from quality issues; thus, cross-validation with national, satellite, and local sources was essential.

Additional constraints arose from the dataset's limited temporal and geographic scope, which may underrepresent long-term trends and obscure region-specific patterns. Techniques such as stratifying data by city and season, along with time-aware cross-validation (e.g., rolling window validation), were employed to preserve local nuances and assess model robustness. Future work should expand the dataset to include more diverse regions and integrate socioeconomic variables—like household income, building insulation quality, and appliance efficiency—to further refine predictions. Moreover, applying interpretability tools such as SHAP could demystify the "black-box" nature of the hybrid model.

From a policy perspective, the enhanced forecasting capability supports several initiatives:

- **Dynamic Demand-Side Management:** Adaptive pricing during peak demand can incentivize reduced usage, alleviating grid strain.
- **Infrastructure Prioritization:** Investments in pipeline resilience and storage should focus on regions where consumption peaks are most severe, such as high-altitude areas.
- **Integration with Renewable Energy:** Hybrid energy systems (e.g., gas-solar) can reduce reliance on single-source heating during winter months.
- **Real-World Implementation:** Incorporating real-time smart meter data and engaging the public through tailored energy conservation campaigns will further enhance the practical impact of predictive models.

In summary, while this study successfully leverages weather data and advanced machine learning to bridge the gap between coarse consumption records and daily climatic variations, addressing data granularity and scope limitations will be critical. Expanding to longitudinal, high-resolution, and socioeconomically enriched datasets will not only improve forecast accuracy but also empower policymakers to develop more resilient, efficient, and equitable gas distribution strategies in climate-vulnerable regions.